\pdfoutput=1

\documentclass[11pt]{article}

\usepackage{ACL2023}

\usepackage{graphicx}
\usepackage{amsmath}
\usepackage{amssymb}
\usepackage{times}
\usepackage{latexsym}
\usepackage{booktabs, multirow} 

\usepackage[T1]{fontenc}

\usepackage[utf8]{inputenc}

\usepackage{microtype}

\usepackage{inconsolata}

\newcommand{\head}[1]{\noindent\textbf{#1}}

\newcommand{\eg}{\textit{e.g.}}

\newcommand{\ct}{ChartT5}
\NewDocumentCommand{\heng}
{ mO{} }{\textcolor{cyan}{\textsuperscript{\textit{Heng}}\textsf{\textbf{\small[#1]}}}}

\title{Enhanced Chart Understanding in Vision and Language Task via Cross-modal Pre-training on Plot Table Pairs}

\author{
Mingyang Zhou$^1$, Yi R. Fung$^2$, Long Chen$^1$, Christopher Thomas$^3$, \\ \textbf{Heng Ji$^2$, Shih-Fu Chang$^1$}  
\\ 
$^1$Columbia University \quad $^2$University of Illinois at Urbana-Champaign  \quad  $^3$Virginia Tech \\
\texttt{\{mz2974, cl3695, sc250\}@columbia.edu, \{yifung2,hengji\}@illinois.edu}, \\ \texttt{chris@cs.vt.edu} \\
}

\begin{document}
\maketitle
\begin{abstract}
Building cross-model intelligence that can understand charts and communicate the salient information hidden behind them is an appealing challenge in the vision and language (V+L) community. The capability to uncover the underlined table data of chart figures is a critical key to automatic chart understanding. We introduce \ct, a V+L model that learns how to interpret table information from chart images via cross-modal pre-training on plot table pairs. Specifically, we propose two novel pre-training objectives: Masked Header Prediction (MHP) and Masked Value Prediction (MVP) to facilitate the model with different skills to interpret the table information. We have conducted extensive experiments on chart question answering and chart summarization to verify the effectiveness of the proposed pre-training strategies. In particular, on the ChartQA benchmark, our \ct~outperforms the state-of-the-art non-pretraining methods by over $8\%$ performance gains.
\end{abstract}

\section{Introduction}
Chart figures serve as the visual summary of tabular data, which helps to convey rich context in various documents, such as scientific papers, textbooks, and technical news. An intelligent agent that can understand and communicate chart plots can lead to many useful applications. For example, a virtual doctor who knows how to answer the patient's question on a complex medical report or a reading assistant who can summarize the key findings from scientific papers in brief language. In the past few years, there has been a growing interest in our community to explore chart understanding in vision and language (V+L) tasks and many related benchmarks like Chart Question Answering (\textbf{CQA})~\cite{2022-chartqa, kafle2018dvqa, Methani_2020_WACV} and Chart Summarization (\textbf{CS})~\cite{2022-chart} are introduced.

While prevalent in the research community, automatic chart understanding remains a challenging problem due to its complex compositions of various shapes, lines, colors, and scene text. Although tremendous success is achieved in the V+L research, applying these existing methods to handle chart-related tasks is hard. Recent research ChartQA~\cite{2022-chartqa} and Chart-to-Text~\cite{2022-chart} attempt to first convert chart images to their underlined tables and use the extracted tables to perform chart-related V+L task. As the extracted tables always have clean and organized structures, it makes extracting relevant information to solve downstream reasoning tasks much more accessible. Empirically, using tables yields promising results on both CQA and CS.

\begin{figure}[t!]
\centering
\includegraphics[width=\linewidth]{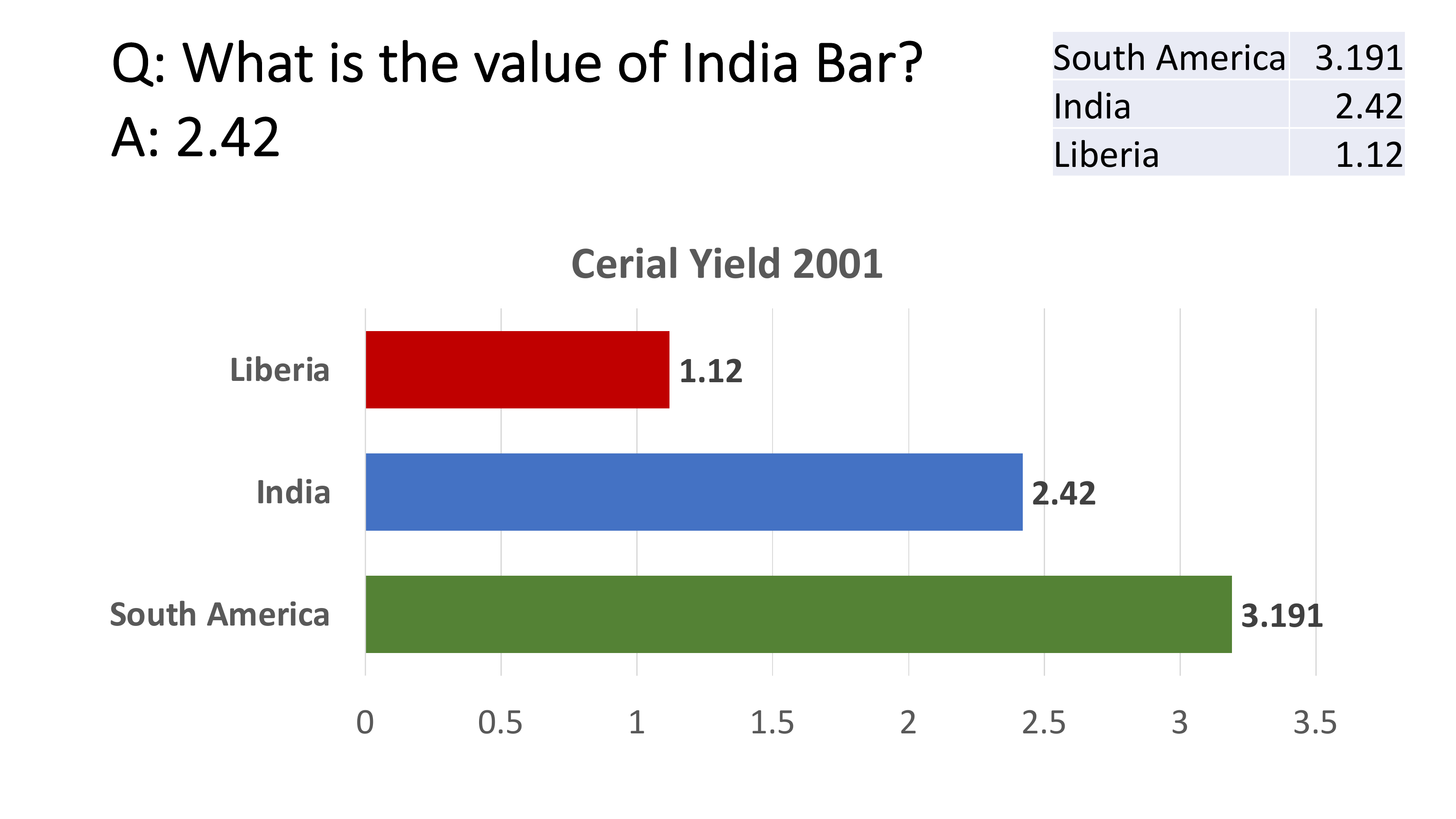}
\caption{A data sample from the ChartQA dataset. The corresponding chart table is displayed in the top right corner. }
\label{fig:Data_Example}
\end{figure}

Despite valuing table as a significant ingredient for chart understanding, we have two main concerns about this approach: (1) Automatic table extraction is unreliable. Existing methods~\cite{luo2021chartocr, parsingline} are often limited to work on a few particular types of chart images and do not generalize well. Moreover, the extracted table is likely to contain incorrect noisy predictions that potentially harm the performance of the following task. (2) In most cases, the whole table is optional for resolving the chart-related V+L task. As illustrated in Fig~\ref{fig:Data_Example}, to answer the question \textit{"What is the value of India Bar"}, the model just needs access to the second row to give the correct answer. In contrast, having redundant table information makes finding the relevant information challenging. To better leverage the table data, we argue that it is important to equip the V+L model with the capability to dynamically interpret the table value from the chart information. 

Therefore, in this paper, we propose \textbf{\ct}, an OCR-based image-to-text generation model pre-trained on a self-collected chart table pairs corpus. More specifically, \ct~learns how to uncover a masked table with two proposed pre-training objectives: Masked Header Prediction (MHP), and Masked Value Prediction (MVP). MHP helps improve the model's capability of linking scene text to the corresponding table headers. MVP requires the model to perform mathematical reasoning over chart structure units and the scene text to predict the correct data value. 

We evaluate our \ct~on two tasks and benchmarks: ChartQA and Chart-to-Text. In ChartQA, \ct~outperforms all the non-pretraining methods that use extracted tables by at least 8$\%$ performance gains. \ct~also beats the pre-training table-based methods, which demonstrates the effectiveness of the proposed pre-training strategies. On Chart-to-Text, \ct~consistly outperforms the existing SOTA on the content selection metrics~\cite{barzilay-2005-collective} which values the model's capability to extract the critical information from the chart. 

In summary, our contributions are summarized below:
\begin{itemize}
\itemsep-0.2em

\item We propose chart-to-table pre-training for V+L model to learn the capability of interpreting table data from the chart.

\item We demonstrate that the pre-trained model consistently outperforms table-based methods on two chart understanding tasks. 

\item We conduct comprehensive ablation studies to validate the effectiveness of chart-to-table pre-training and the proposed pre-training objectives.
\end{itemize}

\begin{figure*}[ht!]
\centering
\includegraphics[width=\linewidth]{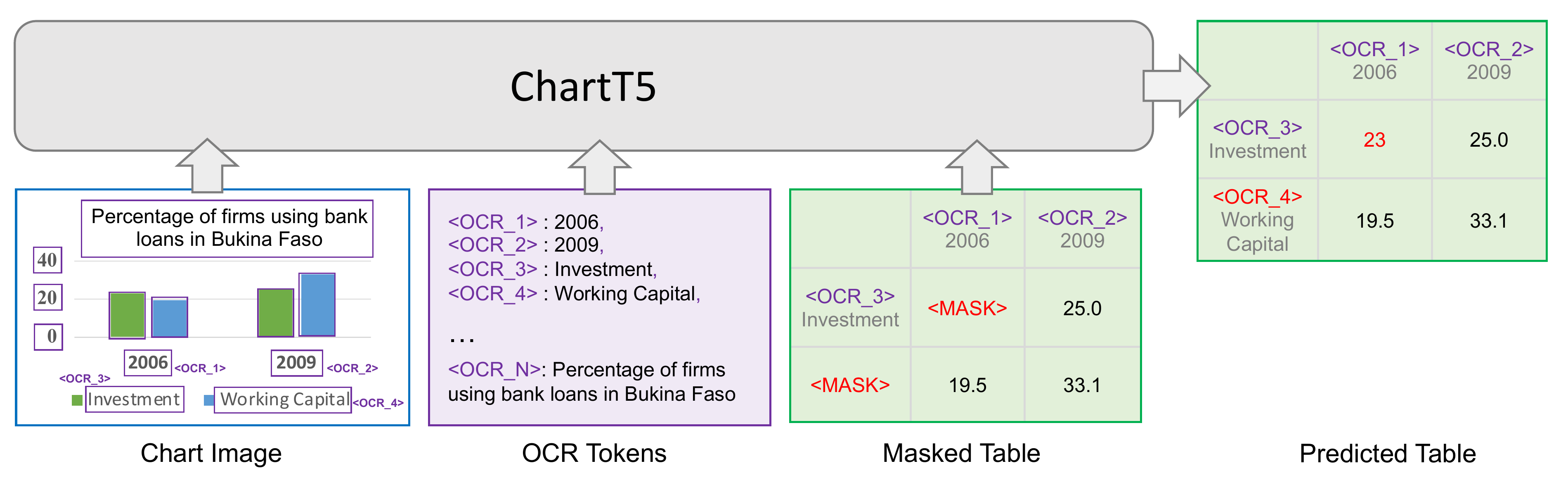}
\caption{An Overview of \ct. Given the input chart image and the extracted OCR tokens, \ct~predicts the masked values of the table in the output. }
\label{fig:model}
\end{figure*}

\section{Related Work}
\subsection{Vision and Language Research on Charts}
Researching chart understanding in V+L tasks is a popular field nowadays. The most prevalent problem is chart question answering (CQA)~\cite{kafle2018dvqa,ebrahimi2018figureqa,Methani_2020_WACV,2022-chartqa,leafqa}, where researchers build models to answer complex questions on chart images. Another popular one is chart summarization (CS)~\cite{2022-chart, obeid-2020-chart}, which requires machine learning models to create a summary of key insights conveyed by a chart. \citet{scicap-generating} collected a large-scale scientific figures captioning dataset from research papers where many images are chart plots. 

There are two main approaches for chart vision and language tasks. The first approach adapts existing visual question answering (VQA) and image captioning models to CQA and CS tasks with some specialized designs for chart images~\cite{kafle2020answering,singh-2020-stl,leafqa,kafle2018dvqa,scicap-generating,Spreafico2020NeuralDC}. The other approach assumes the table data of charts is accessible from the dataset~\cite{Kim2020AnsweringQA, Masry2021IntegratingID} or can be extracted from the chart images using vision to table techniques~\cite{Methani_2020_WACV, 2022-chartqa, 2022-chart}. Then, the researchers will either use a table-to-text generation model~\cite{Kim2020AnsweringQA, Masry2021IntegratingID, Methani_2020_WACV} or combine the embedding of tables and charts via a multi-modal fusion method to generate the text output~\cite{2022-chartqa, 2022-chart}. It is clear from these efforts that adding tables as the additional representation of charts will dramatically improve the model's capability to understand and interpret chart information. 

Following the table-based approach, we also value the information provided by the underlined table data of chart images. However, instead of directly concatenating the extracted table into the chart understanding model, we facilitate our model with the capability to interpret the table data from chart images via pre-training on chart-table pairs.

\subsection{Vision and Language Pre-training}
Vision and language pre-training has received growing interest over the past few years. Researchers build transformer-based multi-modal fusion models and perform self-supervised learning on a large-scale corpus of image-text pairs to learn robust cross-modal representations that can benefit the performance of various downstream tasks \cite{chen2020uniter, lu2019vilbert, tan2019lxmert, su2019vl, li2020oscar,Zhang_2021_CVPR}.

While the pre-trained models achieve great success on tasks like VQA~\cite{antol2015vqa} and Image Captioning~\cite{chen2015microsoft}, they have only focused on the domain of natural images. However, chart understanding is still challenging for the existing vision and language methods due to their lack of knowledge of scene text and structured visual units such as ``bars'' and ``lines''. 

To address the limitation of conventional vision and language pre-training, TAP~\cite{yang2021tap} and PreSTU~\cite{PreSTU} propose OCR-based vision and language pre-training frameworks that focus on scene text understanding in natural images where they design various pre-training objectives around the extracted OCR texts. Most recently, Donut~\cite{kim2022donut} and Pix2Struct~\cite{pix2struct} propose OCR-free pre-training frameworks, where the pre-trained model directly generates a text output from a raw image input. Donut focuses on document image (\eg, receipt) understanding, and Pix2Struct aims to handle broader types of synthetic images that contain visually-situated texts such as infographics and user interfaces via parsing web-page screenshots into their HTML Code. Different from these works, we take the first step to explore vision and language pre-training that focuses on chart image understanding. Specifically, we propose novel pre-training objectives to parse charts to their underlined tables.

\section{Method}
In this section, we first introduce the dataset for pre-training. We then go over our \ct~model architecture and pre-training objectives to predict masked tables from the chart and OCR information. 

\subsection {Pre-training Dataset Collection}

To collect large-scale pairs of chart-table data, we collect synthetic data from existing chart question-answering corpora, including PlotQA~\cite{Methani_2020_WACV}, DVQA~\cite{kafle2018dvqa}, and FigureQA~\cite{ebrahimi2018figureqa}. Specifically, DVQA and FigureQA render chart images from synthetic tables that are randomly generated from limited vocabularies. PlotQA first scrapes tables from online resources like World Bank Open Data and then synthesizes the charts from the scraped data, where the tables and charts contain more diverse language information. Our pre-training corpus consists of 495K chart-table pairs, which cover a diverse range of chart types. Our pre-training corpus contains three chart types: bar, line, and pie. The distribution of different chart types from the three chart question-answering benchmarks is summarized in table \ref{tab:data_stats}.

\begin{table}[t]
\centering
\small
\begin{tabular}{l|ccc|c}\toprule
Type & PlotQA & DVQA & FigureQA & Total  \\\cmidrule{1-5}
Bar  & 142,587 & 204,514  & 40,000 & 387,101\\
Line & 48,133 & 0 & 40,000 & 88,133 \\
Pie & 0 & 0 & 20,001 & 20,001 \\
\bottomrule
\end{tabular}
\caption{Distribution of the three chart types: bar, line, and pie from different resources in the pre-training corpus.}
\label{tab:data_stats}
\end{table}


\subsection{Model Overview}
ChartT5 is an extension of the existing V+L Pre-training framework, VLT5 \cite{cho2021vlt5}, an encoder-decoder architecture that unifies the vision-language tasks as text generation conditioned on multi-modal inputs. 
Given a chart image, we first extract the scene texts. For the synthetic chart images that are collected from DVQA~\cite{kafle2018dvqa}, FigureQA~\cite{ebrahimi2018figureqa}, and PlotQA~\cite{Methani_2020_WACV}, the ground-truth scene texts are available. The visual context is then represented as combining visual features extracted from the chart image and the language features obtained on the detected scene text. We then flat the paired table of the chart image into a string and extract the text features via the language encoder. The multi-modal features are then concatenated and fused via the multi-layer encoder, and the output hidden vectors can then be used for various pre-training tasks. 

\subsubsection{Chart Image Encoder}
Given an input chart image, to recognize the critical marks (\eg, bars and lines) of chart images, we first utilize a pre-trained Mask R-CNN object detector from~\cite{2022-chartqa} to extract the visual region features $\boldsymbol{v} = \{v_1, v_2, \cdots, v_{l^v}\}$. Next, the chart object detector is trained on the synthetic chart images from the previous CQA datasets \cite{ebrahimi2018figureqa, kafle2018dvqa, 2022-chartqa, Methani_2020_WACV} which is defined to identify 15 chart-related objects\footnote{These 15 categories are: Legends, yAxisTitle, ChartTitle, xAxisTitle, LegendPreview, PlotArea, yAxisLabel, xAxisLabel, LegendLabel, PieLabel, bar, pie, pieSlice, line, and dotLine.}. For each detected object region, we also extract location features as a 5-d vector: [$\frac{x_1}{W}$,$\frac{y_1}{H}$,$\frac{x_2}{W}$,$\frac{y_2}{H}$,$\frac{(y_2-y_1)(x_2-x_1)}{W.H}$], which denotes the normalized top left coordinates, bottom right coordinates, and the normalized area of the detected region box. The position feature is then fed through fully-connected layers to be projected to the visual region feature embedding space. The final representation of the visual feature is obtained by summing up the projected region feature and corresponding location feature. 

\subsubsection{OCR Encoder}
After extracting the list of the OCR words from the chart image, we obtain a set of OCR text embeddings $\boldsymbol{o} = \{o_1, o_2, \cdots, o_{l^o}\}$ via a learned word embedding layer. We also get each OCR token's 5-d position vector similar to the visual position vector from the OCR token's detected bounding box. We then obtain the position embedding vector using the shared projecting layer from the Chart Image Encoder. The shared position encoding mechanism between OCR tokens and chart object regions would help the model to capture their relative positional relations, which is a critical clue to predict the table data from the chart image. For example, the bar associated with an x-axis label should share a similar x-coordinate position in a vertical bar chart. The final OCR embedding vector is gained by summing up the OCR text token embeddings and the OCR position embedding. 

\subsubsection{Language Encoder}
Following the setting of the original VLT5~\cite{cho2021vlt5}, we add a prefix to the flattened underlying table to indicate different pre-training tasks. We then get the table token embeddings $\boldsymbol{t} = \{t_1, t_2, \cdots, t_{l^t}\}$ with a shared word embedding layer. We apply the original T5's \cite{2020t5} relative position bias to obtain the position information of each token in the caption and the flattened table. We know that the tables have very different structures compared to natural language captions, and several efforts are exploring specialized position embeddings for tables~\cite{yin20acl, herzi-2020-tapas}. We leave the exploration of the specialized table position embedding for chart table pre-training in the future. 

\head{Scene Text Copy Mechanism.} A critical ingredient to the success of chart-to-table translation is the ability to predict the table headers from the corresponding OCR texts. For example, in the horizontal bar chart, the table column header is usually obtained from the x-axis labels, and the row header is often copied from the legend labels. Although presenting OCR text and the table to the model helps link the shared OCR tokens and table values, generating the correct table prediction from the corresponding OCR source is still challenging due to the large candidate token vocabulary. 
To encourage direct copy from the OCR text to the associated table cell value, we introduce OCR sentinel tokens $\{<\text{ocr}\textunderscore1>, <\text{ocr}\textunderscore2>, \cdots, <\text{ocr}\textunderscore{l^o}> \}$, which corresponds to the detected OCR texts.
As illustrated in Figure~\ref{fig:model}, we replace each OCR token with a unique corresponding OCR sentinel token. Then, for every OCR token, we find if there is a matched existing table cell value. If a matched pair is found, we replace the table cell value with its paired OCR sentinel token. During pre-training, as all the plot images are synthesized from a paired table, the one-to-one scene text to table value mapping is already provided. With this prepossessing procedure, we successfully distinguish the table values that are copied from OCR tokens and those that need to be generated from the general token vocabularies, encouraging more accurate table prediction from the relevant resources. 


\subsection{Pre-training Objectives}
Given the chart-table pairs, we propose Masked Header Prediction (MHP) and Masked Value Prediction (MHP) to teach the model to recover incomplete tables with the chart information. Specifically, this objective aims to predict a masked table token $t_m$ with the remaining table info $t_{\backslash m}$ as well as the chart image region $\boldsymbol{v}$ and the scene text $\boldsymbol{o}$. Compared to the traditional masked language modeling applied to the natural language text, we adjust the table masking strategy based on two hypotheses: (1) We alternatively mask just the table headers or numerical table values, as we think interpreting these two types of information requires different skills. Predicting table headers requires retrieving the correct scene text, while predicting numerical table values depends more on the capability to conduct mathematic reasoning over both the visual elements and the scene text. Therefore, it is better to format them as two separate pre-training objectives. (2) We increase the masking rate from 15$\%$ to 45$\%$, as the masked table token has less dependence on the surrounding table values.


\section{Experiment}
In this section, We detailed our experiment setups to evaluate the proposed \ct~on two tasks: chart question answering and chart summarization. We then introduce the main results of the two evaluation tasks. Finally, we present the ablation study on chart-table pre-training and the two pre-training objectives.

\head{Chart Question Answering.} Given a chart image and a query question, the goal for the model is to provide an accurate answer string by interpreting the provided chart image. For this task, we consider the ChartQA dataset \cite{2022-chartqa}, which collects question-answer pairs on realistic chart images scraped from the internet. Their annotations are collected in two fashions: (1) Human-written question-answer pairs; and (2) machine-generated question-answer pairs derived from the human-written chart summaries. In total 32.7K question-answer pairs are collected on 21.9K scraped chart images, where about 9.6K question-and-answer pairs are human-written. Compared to the previously collected CQA datasets, ChartQA is more challenging to handle due to the diverse visual style from the realistic chart images and the complex language from human annotations. Following previous work \cite{2022-chartqa, Methani_2020_WACV}, we also apply the relaxed accuracy to measure the performance on the CQA task, which allows a minor inaccuracy on numerical value prediction (within 5$\%$ of the gold answer). For non-numerical answers, the prediction needs to be exactly matched to the gold-standard answer. 

\head{Chart Summarization.} Given a chart image, the target is to summarize the key insights of the chart in natural language. For this task, we evaluate our model on the most recently proposed Chart-to-Text benchmark~\cite{2022-chart}, which collects roughly 36.5K chart images with one summary for each image. They split the collected charts into two sets: Statista and Pew, representing the two separate websites from which the chart plots come. The summaries in Statista are human-written which is well grounded on the chart image.
Meanwhile, the summaries from Pew are automatically extracted from the news paragraphs surrounding the chart images. Pew is noisier and more challenging to handle. We follow \cite{2022-chart} to split the two sets for training and testing. We adopt BLEU-4, Content Selection, and CIDER as the evaluation metrics to measure the quality of the generated summary following \cite{2022-chart}.  

\head{Implementation details.}
We initialized our \ct~from $\text{T5}_{\text{base}}$ and pre-trained on our self-collected corpus for 30 epochs with a batch size of 60. We used Adam optimizer~\cite{ADAM} with a linear warm-up for the first 5$\%$ training steps, and the peak learning rate is set as 1e-4. After warming up, a linear decayed learning-rate scheduler gradually drops the learning rate for the rest of the training steps. The pre-training experiments are conducted on 2 Nvidia TITAN RTX GPUs, and it roughly takes two days to accomplish the experiment. We kept the last checkpoint of each pre-training run as our final checkpoint for fine-tuning. 

We also applied warming-up for downstream fine-tuning to gradually increase the learning rate to the pick value during the first 5$\%$ of training epochs. After that, a linear decayed learning-rate scheduler gradually drops the learning rate for the remaining training. For CQA task, we set batch size as 24 and fine-tune \ct~for 60 epochs with a peak learning rate 2e-4 on 2 Nvidia TITAN RTX GPUs. The best checkpoint was saved as the one that achieves the highest accuracy on the validation split. On the CS task, we use batch size 20 and a peak learning rate 5e-5. On the Pew split, we fine-tune \ct for 20 epochs, and on Statista, we fine-tune \ct for 25 epochs. The best checkpoint is also saved as achieving the best BLEU score on the validation split. All the reported numbers are one-time runs.  


\subsection{Main Results}
We first compare ChartT5 to various state-of-the-art methods with or without pre-training on the two downstream tasks. 

\begin{table}[t]
\centering
\begin{tabular}{l|ccc}\toprule
\multirow{2}{*}{ Model } & \multicolumn{3}{c}{ChartQA} \\
&Human &Augment & Overall \\\cmidrule{1-4}
T5 & 25.12  & 56.96 & 41.56\\
Tapas & 28.72 & 53.84 & 41.28 \\
VLT5 &26.24 & 56.88  &41.56  \\
VisionTapas & 29.60& 61.44 & 45.52 \\
\midrule
$\text{VLT5}_{pre}$ & \textbf{40.08} &63.60 & 51.84 \\
$\text{VisionTapas}_{pre}$ & 32.56 & 61.60 &47.08 \\
Pix2Struct & - & \ - & \textbf{56.00}  \\
ChartT5 & 31.8 & \textbf{74.4} &53.16 \\
\bottomrule
\end{tabular}
\caption{Evaluation results on ChartQA. We report relaxed accuracy on the test split annotated by humans and that generated by the machine. In the last column, we report the overall accuracy by computing the mean values with human split and augment split.  
}
\label{tab:main_chartQA}
\end{table}


\subsubsection{Evaluation on CQA}
We compare ChartT5 with SOTA non-pretraining and pre-training methods on CQA tasks. The best-performed non-pretraining baselines are introduced in \cite{2022-chartqa}. The authors first predict the table data from the chart image via an automatic data extraction tool \cite{luo2021chartocr}. Then they extend various language-only models (T5, Tapas) and multi-modal models (VLT5, VisionTapas) to predict the answer conditioned on the extracted table. On the line of pre-training baselines, we compare to $\text{VLT5}_{pre}$ and $\text{VisionTapas}_{pre}$ which pre-trains VLT5 and Vision Tapas on PlotQA with the visual question answering tasks. We also compare chartT5 to the current SOTA method Pix2Struct which is pre-trained on 80 million webpage screenshots to HTML code parsing objectives. The result is summarized in Table~\ref{tab:main_chartQA}. 

\head{Comparison to Non-Pretraining Method}
Even without access to the predicted tables, \ct~has outperformed all non-pretraining methods by a large margin (a minimum 7.3$\%$ gain on the overall performance). \ct~also outperforms all non-pretraining baselines on the human-written questions and machine-generated questions. Although the predicted table covers 54$\%$ of the answers in the test data of ChartQA, simply feeding it as an input does not make the existing models fully leverage the valuable information. The significant improvement achieved by \ct~indicates the effectiveness of the proposed pre-training to help the model to obtain the relevant table information for chart understanding. 

\head{Comparison to Pre-training Method}
Although the performance of VLT5 and VisionTapas is improved significantly by pre-training on additional CQA data, ChartT5 still outperform them by at least 1.3$\%$. Specifically, on machine-augmented questions, ChartT5 outperforms $\text{VLT5}_{pre}$ by 8$\%$. However, both $\text{visionTapas}_{pre}$ and $\text{VLT5}_{pre}$ achieve better accuracy on the human split, which means that the in-domain question answering objectives helps the model to improve the numerical reasoning capability. 
ChartT5 underperforms Pix2Struct by 2.3$\%$ on the overall test split. However, pix2struct is pre-trained on a more than 100 times larger pre-training corpus than the rest of the pre-training methods. Given the same scale of the pre-training dataset, we expect to gain additional performance improvement, and we leave this for future exploration.

\begin{table*}[t]\centering
\begin{tabular}{l|ccc|ccc}\toprule
\multirow{2}{*}{ Model } & \multicolumn{3}{c|}{Statista} & \multicolumn{3}{c}{Pew} \\
&BLEU &CS & CIDER & BLEU &CS & CIDER \\\cmidrule{1-7}
T5-OCR & 35.29 & 73.77 & 4.43 & \textbf{10.49}  &40.87 & \textbf{2.20}  \\
BART-OCR & - & - & - & 9.09 & 39.99 & 1.97 \\
T5-TAB & 37.01 & 75.72 & \textbf{4.68} &  - & - & -\\
BART-TAB & 36.36 & 77.14 & 4.40 & - & - & - \\ 
ChartT5 & \textbf{37.51} & \textbf{82.16} & 3.45 & 9.05 & \textbf{55.1} & 1.23\\
\bottomrule
\end{tabular}
\caption{Evaluation results on Chart Summarization. We display BLEU, CS and CIDER scores for the Pew and Statista Split. The ground truth table is not available to Pew thus the table-based method does not have results on Pew split.}
\label{tab:main_chartSum}
\end{table*}

\subsubsection{Evaluation on Chart Summarization}
For the chart summarization task, we compare \ct~to the best non-pretraining approaches introduced in \cite{2022-chart}. Given a chart image, The authors build the chart summarization models by extending the pre-trained language generation model T5~\cite{2020t5} and BART\cite{Lewis2019BARTDS} whose generation processes are conditioned on: (1) a set of scene texts extracted by a trained OCR detector. (2) the ground truth table that is paired with the chart. The evaluation result is summarized in Table \ref{tab:main_chartSum}.

From Table \ref{tab:main_chartSum}, we can see that on Statista, ChartT5 outperforms all baseline methods on BLUE score, but only a slight improvement is achieved over the best baseline. On Pew, ChartT5 underperforms T5-OCR by almost 1.5 percent. The proposed ChartT5 also slightly underperforms against the baseline methods in CIDER on both datasets. However, \ct~consistently outperforms all baselines on content selection scores across both Statista and Pew sets. 
The under-performance on BLEU and CIDER indicates that Chart-table pre-training is limited to benefit high-quality natural language generation. However, the strong performance on content selection, which values the key information appearance in the generation, suggests the advantage of chart-table pre-training on extracting relevant chart information. Therefore, a potential direction to explore is combining different types of pre-training objectives, such as chart-to-text pre-training and chart-table pre-training goals, to facilitate the model with diverse strengths. 

\begin{table}[t]
\centering
\begin{tabular}{l|ccc}\toprule
\multirow{2}{*}{ Pretraining? } & \multicolumn{3}{c}{Question Types} \\
&Table & Human & Augment  \\\cmidrule{1-4}
No  & 60.7 & 30.8  & 66.7\\
Yes & \textbf{64.7} & \textbf{31.8} & \textbf{74.4} \\
\bottomrule
\end{tabular}
\caption{Ablation Study on Chart Table Pre-training with ChartQA Dataset. We report results on three subsets of questions: Table cover questions, human-written questions, and machine-generated questions. }
\label{tab:main_chartQA_ablation_pretrain}
\end{table}

\subsection{Ablation Study}
We conduct ablation experiments to validate the effectiveness of chart-table pre-training and the pre-training objectives. We also evaluate the effectiveness of the proposed scene text copy mechanism. 
\subsubsection{Chart-Table Pre-training} We conduct detailed analyses on the effectiveness of chart-table pre-training. First, we measure the performance gain from the chart-table pre-training on the full test set of ChartQA data. We then study what type of questions benefit most from the chart-table pre-training by picking three subsets of questions that measure different capabilities of the model: (1) Human-written questions, (2) Machine-generated questions, and (3) Table covered questions, where the answers can be directly found in the ground truth tables. The results are summarized in Table \ref{tab:main_chartQA_ablation_pretrain}. From Table \ref{tab:main_chartQA_ablation_pretrain}, we find that after chart-table pre-training the model's performance on these three sets of questions is all improved. The most significant gain is obtained on machine-generated questions, which mainly focus on extractive-type questions. This indicates that chart-table pre-training benefits the model to localize and retrieve the requested information presented on Chart Image. The second biggest gain is achieved on table-cover questions, where the model demonstrates significant improvement in the capability of chart-to-table interpretation. 

\begin{table}[!ht]
\centering
\begin{tabular}{l|ccc}\toprule
\multirow{2}{*} & \multicolumn{3}{c}{Question Types} \\
& Human & Augment & Overall \\\cmidrule{1-4}
Full  & 31.8  & 74.4 & 53.1\\
- MVP & 30.9 & 73.7  & 52.3 \\
- MHP & 31.2 &  68.3 &  49.7 \\
- STC & 30.8 & 72.4 & 51.6 \\
\bottomrule
\end{tabular}
\caption{Ablation Study on the two proposed pre-training objectives and the Scene Text Copy Mechanism (STC). The first row is the result of the full \ct~model. Then we remove one of the pre-training objectives and the scene-text-copy mechanism. We report the results of different ablation experiments on both human and machine-generated splits as well as the overall performance.}
\label{tab:main_chartQA_ablation_pretrain_obj}
\end{table}

\begin{figure}[t!]
\centering
\includegraphics[width=\linewidth]{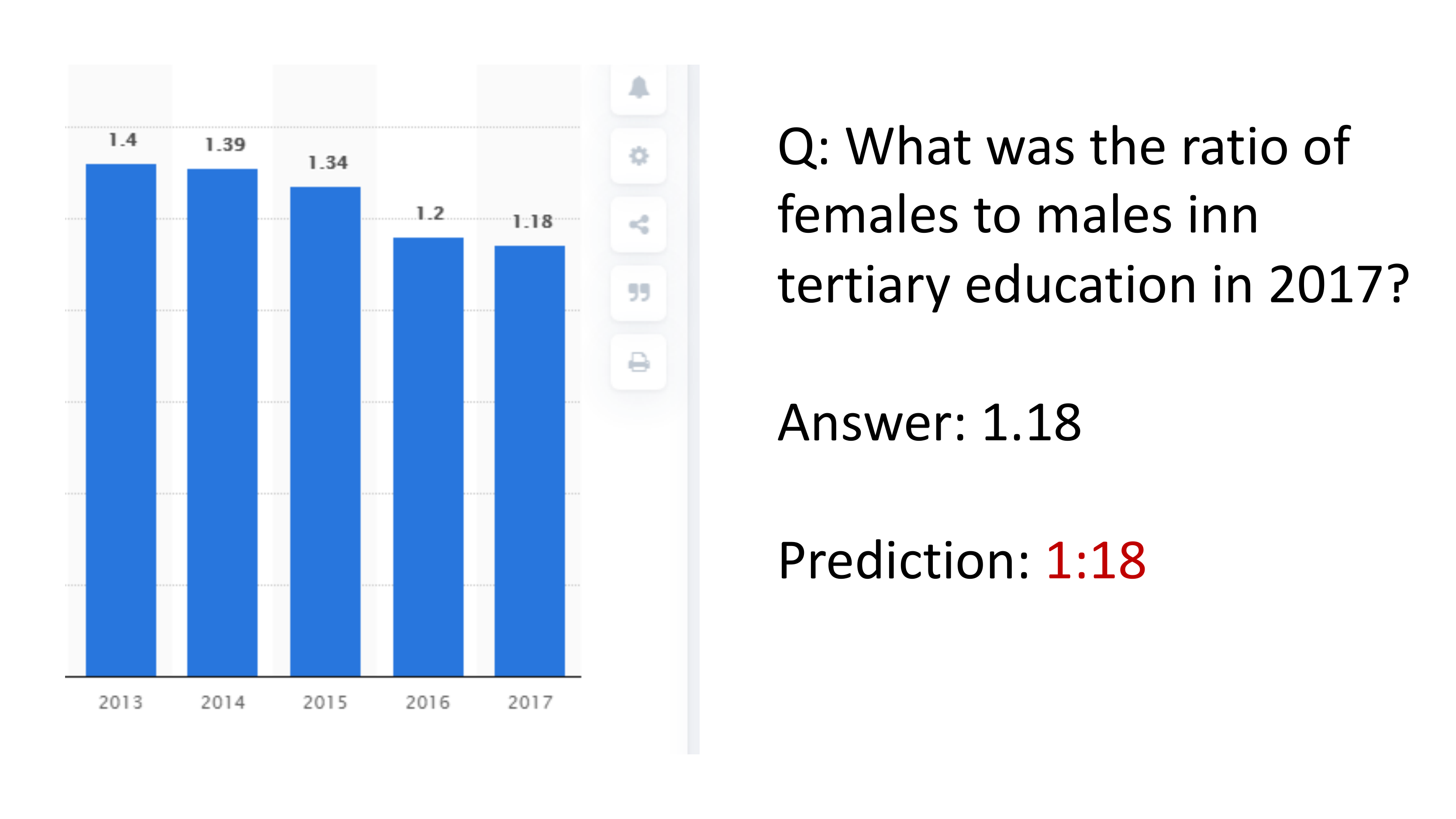}
\caption{An error prediction from our model due to noisy OCR prediction}
\label{fig:Error_Example}
\end{figure}

\begin{figure}[t!]
\centering
\includegraphics[width=\linewidth]{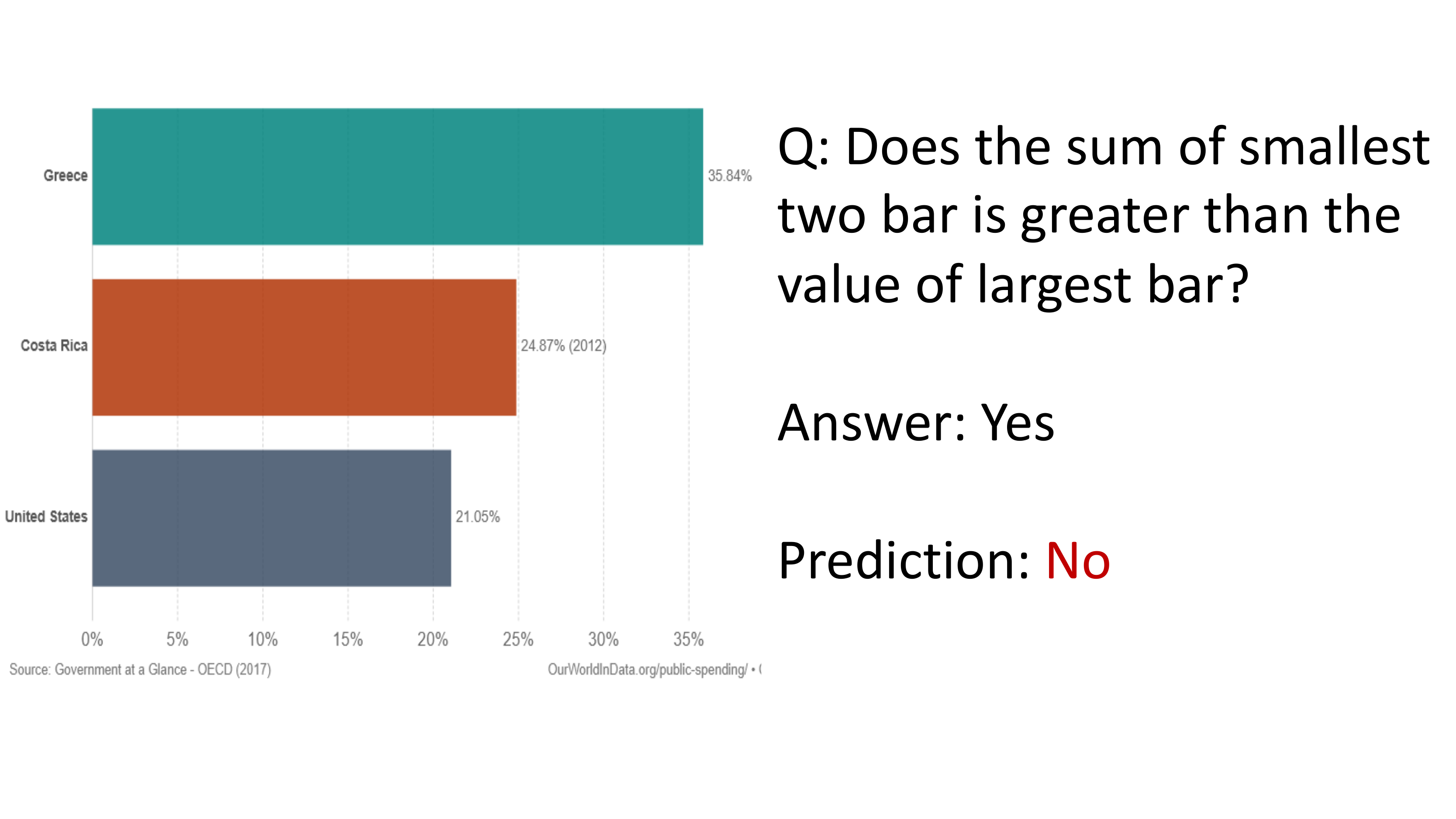}
\caption{An error prediction from our model due to complex multi-hop reasoning}
\label{fig:Error_Example_2}
\end{figure}

\subsubsection{Pre-training Objectives} We validate the effectiveness of the two pre-training objectives, Masked Header Prediction and Masked Value Prediction. We remove one pre-training objective at a time and pre-train the \ct with only one table prediction task. The pre-trained model is then fine-tuned and evaluated on the human and augmented split for comparison. The result is displayed in table \ref{tab:main_chartQA_ablation_pretrain_obj}. As can be seen from the table, removing Masked Value Prediction Loss has a negligible impact on the performance of \ct~on ChartQA dataset. There is a slightly more drop in human written questions which suggests that predicting table numerical values still has a miner positive impact on helping the model's mathematical reasoning. Remove Masked Header Prediction have a significant impact on the machine-generated question-answering accuracy. As expected, Masked header modeling mainly helps the model learn how to link the scene text to the table headers, which is a critical ability to extract relevant information given a specific query.  

\subsubsection{Scene Text Copy}
We also validate the effectiveness of the scene-text-copy mechanism, where we train a \ct model by simply representing OCR tokens in their original text format. The model is fine-tuned and evaluated on the human and augmented split of the chartQA dataset to compare against the full \ct. The result is displayed in Table \ref{tab:main_chartQA_ablation_pretrain_obj}. Disabling the scene-text-copy mechanism leads to a 1.5$\%$ overall performance drop on ChartQA tasks. Specifically, it leads to more degradation on the augmented split than the human split, as scene-text-copy helps enhance the alignment between OCR and table values to benefit accurate information extraction from the chart. 

\subsection {Qualitative Error Analysis}
We have manually analyzed model predictions to understand its limitation. We found that our model suffers most from noisy OCR detection and complex question that requires multi-hop reasoning. 

\head{Noisy OCR Prediction.} As an OCR-based model, \ct~often suffers from a wrong OCR detection. An example is shown in Figure~\ref{fig:Error_Example}; the model localizes the right scene text ``1.18'' to answer the question, but the OCR text is mistakenly detected as ``1:18''. To further understand the limitation of OCR detection, we randomly sample 20K PlotQA test split and compare the performance of our model using detected OCRs against Ground Truth OCRs. We observe a 5\% performance drop when using detected OCRs. We can improve the OCR detector for future work by training on a large Plot scene-text detection benchmark. Another promising direction is to attempt OCR-free end-to-end plot recognition method like Pix2Struct \cite{pix2struct}.

\head{Multi-Hop Reasoning.} Our model is also quite vulnerable to handling complex questions requiring multi-hop reasoning. An example is shown in Figure~\ref{fig:Error_Example_2}; the model cannot perform the complex logic reasoning to add the stats of the two smallest bars and compare that to the large bar. We will consider exploring pre-training on the mathematic reasoning datasets to address this limitation.

\section{Conclusion}
We propose \ct~to enhance the vision language model's ability to understand chart images via chart-table pre-training. The model learns to interpret the masked tables via our proposed masked header prediction and masked value prediction objectives.  \ct~achieves significant improvement over table-based non-pretraining SOTA methods on the ChartQA dataset, especially on the extractive question sets. We also achieve a new SOTA Content Selection Score on the Chart-to-text summarization dataset. We conduct comprehensive ablation studies to identify the impact of chart-table pre-training, and we find that the proposed pre-training is extremely helpful to extract accurate information from the Chart. For future research directions, we believe it may also be meaningful to explore chart understanding under data-efficient settings \cite{2022-degree,zeng2023socratic} and for evidence retrieval tasks \cite{lu2022multihop,ji2023deep}.

\section{Limitations}
Although introducing chart value prediction objective, it only provides minor improvement to the model's performance on doing complex reasoning. There is still a large room to improve the model's capability in math calculation. Our model also suffers from the noisy OCR prediction of off-the-shelf object detector, whose performance will depend highly on the extracted OCR text qualities. 
Another possible limitation of our approach is the quality of the pre-training data, which only contains synthetic images. Although the proposed model works fairly well on the ChartQA dataset, it is unclear if the improved performance can be generalized to other realistic chart images. 

\section{Ethics Statement}
When we collect the pre-training dataset, we ensure we respect the intellectual property of dataset sources. All the ChartQA dataset we used for the collection of chart-table pairs allows public access for research. To ensure the reproducibility of our experiment results, we provide details of the hyperparameter setting in our paper, and we will also publish our code later. Our models can mislead the public's understanding of chart content due to the potential bias from our training corpus. Therefore, we don't recommend using our model for any real-world decision on chart images. 

\section*{Acknowledgement}
This research work is supported by U.S DARPA SemaFor Program No. HR001120C0123. The views and conclusions contained in this work only belong to the authors and should not represent the official policies implied by DARPA or the U.S. Government. The U.S. Government is authorized to reproduce and distribute reprints for governmental purposes notwithstanding any copyright annotation therein. We also thank Ahmed Masry and Shankar Kantharaj for providing us with ChartQA and Chart Summary-related data and baseline model outputs.

\newpage
\bibliography{anthology,custom}
\bibliographystyle{acl_natbib}




\end{document}